\documentclass{article}

\usepackage{arxiv}          
\usepackage[T1]{fontenc}    
\usepackage{hyperref}       
\usepackage{url}            
\usepackage{booktabs}       
\usepackage{amsfonts}       
\usepackage{amsthm}         
\usepackage{nicefrac}       
\usepackage{microtype}      
\usepackage{graphicx}	      
\usepackage{enumitem}	      
\usepackage{tabularx}       
\usepackage{multirow}       
\usepackage{caption}        
\usepackage{siunitx}        
\usepackage[lofdepth,lotdepth]{subfig}
\usepackage[style=numeric,sorting=none]{biblatex}

\date{} 
\renewcommand{\undertitle}{} 
\renewcommand{\headeright}{} 

\newcolumntype{L}{>{\raggedright\arraybackslash}X}
\newcolumntype{C}{>{\centering\arraybackslash}X}
\newcolumntype{R}{>{\raggedleft\arraybackslash}X}
\newcolumntype{O}[1]{>{\raggedright\arraybackslash}p{#1}}
\newcolumntype{P}[1]{>{\centering\arraybackslash}p{#1}}
\newcolumntype{Q}[1]{>{\raggedleft\arraybackslash}p{#1}}

\theoremstyle{definition}
\newtheorem{definition}{Definition}

\hypersetup{
    colorlinks=true,
    urlcolor=blue,
    pdftitle={Class Density and Dataset Quality in High-Dimensional, Unstructured Data},
    pdfpagemode=FullScreen,
}

\title{Class Density and Dataset Quality in High-Dimensional, Unstructured Data}

\author{
  Adam~Byerly\\
  Department of Electronic and Electrical Engineering\\
  Brunel University London\\
  Uxbridge, UB8 3PH UK \\
  Department of Computer Science and Information Systems\\
  Bradley University\\
  Peoria, Il, 61615 USA\\
  \texttt{abyerly@fsmail.bradley.edu} \\
  \And{}
  Tatiana~Kalganova \\
  Department of Electronic and Electrical Engineering\\
  Brunel University London\\
  Uxbridge, UB8 3PH UK \\
  \texttt{tatiana.kalganova@brunel.ac.uk} \\
}

\begin{document}

\maketitle

\begin{abstract}
We provide a definition for class density that can be used to measure the aggregate similarity of the samples within each of the classes in a high-dimensional, unstructured dataset.  We then put forth several candidate methods for calculating class density and analyze the correlation between the values each method produces with the corresponding individual class test accuracies achieved on a trained model.  Additionally, we propose a definition for dataset quality for high-dimensional, unstructured data and show that those datasets that met a certain quality threshold (experimentally demonstrated to be \(> 10\) for the datasets studied) were candidates for eliding redundant data based on the individual class densities.
\end{abstract}


\section{Introduction and Related Work}\label{sec:introduction}

Data density is of interest in the classification domain mostly in cases where there exists a class imbalance problem, which can be quite common in many real-world data collection scenarios (for example credit card fraud detection~\cite{Abdallah2016}\cite{Li2021}).  In this context, the interest is in increasing the density of the data in the minority class via synthetic sample creation~\cite{Chawla2002} or in under-sampling the majority class~\cite{Bunkhumpornpat2017}.  In~\cite{Han2005} and~\cite{Pengfei2014}, the authors showed that generating synthetic minority class samples from data near the boundaries of the classes produced superior results as compared to completely randomly sampling from the entire minority class.  In the clustering domain, data density has been analyzed to produce more intelligent clustering algorithms that do not require the prior selection of the number of clusters (as is required for k-means clustering)~\cite{Hyde2014}.

Studies in data density are replete in the literature for low-dimensional data, such as in those cited in the prior paragraph.  However, data density is less studied in high-dimensional data.  This is most likely due to the fact that distance measurements in high-dimensional space produce near uniform distances~\cite{Aggarwal2001}, a phenomenon colloquially referred to as the \textit{curse of dimensionality}.  As in~\cite{Byerly2022}, in order to reclaim meaningful differences when measuring the distances between samples, in the studies in this work, we use Uniform Manifold Projection and Approximation (UMAP)~\cite{McInnes2018} to reduce high-dimensional image data to 3 dimensions.

In this work, we will be defining a density measurement that can be applied to the 3-dimensional reduction of each class in a dataset independently and then providing a study of classification accuracy after reducing the data in each class to a variety of target densities as opposed to reducing each class by the same amount as in~\cite{Byerly2022}.

As has been put forth and stands to reason, data quality is context dependent~\cite{Bertossi2011}.  Most of the context in the existing literature is on highly-structured, often relational, data for which there are known domains and constraints for well defined groups (tuples) of atomic values~\cite{Bertossi2020}\cite{Borek2011}.
Further, There is no universally agreed upon definition of data quality, let alone a universal method for measuring it.  However, there is widespread agreement on some things, such as having high levels of the following attributes:  timeliness, accuracy, consistency, and completeness~\cite{Blake2011}.  In the present study, we will limit our focus to quality in the sense of completeness.  Only a dataset comprised of low-dimensional data with very limited domains could ever be considered complete in the most strict definition of the term.  For the high-dimensional, unstructured data we will be evaluating, completeness will be demonstrated by removing data from the dataset, based on a target density threshold, while maintaining classification accuracy.

\subsection{Our Contribution}
Our contribution is as follows:
\begin{enumerate}
  \item We provide a definition and method for calculating individual class densities based on 3-dimensional reductions produced by UMAP.\@
  \item We provide a method for calculating an unstructured, high-dimensional dataset's quality, with quality specifically referring to a level of demonstrable completeness.
  \item We show that for 6 out of the 7 datasets studied, datasets with a quality of \(> 10\) could be reduced to a target density of at most 1.0 and achieve accuracy that is statistically insignificantly different from the baseline.  Unique to the dataset for which this was not true (Imagenette) is the relatively low number of training samples.  This combined with the large image size and the assumptions made by UMAP suggest that there is a lower bound on the number of training samples such that any elision whatsoever risks underfitting in the training.
\end{enumerate}

\section{Defining Density}\label{sec:defining_density}

\begin{definition}[Class Density]
  The \textit{class density} of a class is a measure of the aggregate similarity of the training samples available for that class.
\end{definition}

Three candidates for the calculation of a class's density were initially considered, all of them based on the distribution of points in each dimension of any \(m\)-dimensional reduction.  The three candidates are based on the minimum (see \autoref{equation:density_candidate_1}), the maximum (see \autoref{equation:density_candidate_2}), and the mean (see \autoref{equation:density_candidate_3}) standard deviation of the \(m\) gaussians of the \(m\)-dimensional reduction.

For all three candidate calculations, the density is calculated for class \(i\) among all \(n\) classes where \(c_i\) is the count of samples for class \(i\).  \(\sigma_i\) are the standard deviations of the \(m\) gaussians of the \(m\)-dimensional reduction for class \(i\).  

The first three terms of each calculation are identical and are used for biasing unbalanced datasets.  For unbalanced datasets, these terms create a value \(< 1.0\) for those classes that have fewer samples than average and \(> 1.0\) for those classes that have more samples than average.  The motivation for this is to incorporate over and underrepresentation of classes into the density definition.  For balanced datasets, these three terms will compute to 1.0 and thus have no effect on the broader density calculation.

\captionsetup[figure]{list=no, labelformat=empty}

\begin{figure}[!htbp]
    \centering
    \begin{equation}
        d_i^{\,\min} = n \cdot c_i \cdot {\left(\sum^n_{j}c_j\right)}^{-1}
            \cdot \min{\left(\sigma_i\right)}^{-1}
        \label{equation:density_candidate_1}
    \end{equation}
    \caption{\textit{Min-Derived} Density Calculation}
\end{figure}

\begin{figure}[!htbp]
    \centering
    \begin{equation}
        d_i^{\,\max} = n \cdot c_i \cdot {\left(\sum^n_{j}c_j\right)}^{-1}
            \cdot \max{\left(\sigma_i\right)}^{-1}
        \label{equation:density_candidate_2}
    \end{equation}
    \caption{\textit{Max-Derived} Density Calculation}
\end{figure}

\begin{figure}[!htbp]
    \centering
    \begin{equation}
        d_i = n \cdot c_i \cdot {\left(\sum^n_{j}c_j\right)}^{-1}
            \cdot {\left(\frac{1}{m}\sum^m_{k}\sigma_{i_{k}}\right)}^{-1}
        \label{equation:density_candidate_3}
    \end{equation}
    \caption{\textit{Mean-Derived} Density Calculation}
\end{figure}

The {densities} were calculated using each of these calculations for each of the classes for MNIST~\cite{Lecun2010}, Fashion-MNIST~\cite{Xiao2017}, CIFAR-10~\cite{Krizhevsky2009}, CIFAR-100~\cite{Krizhevsky2009}, and Imagenette~\cite{imagenette}.  Then the individual class accuracies for every class in each of these datasets were averaged across the five trials of the baseline experiments.  Then correlations were calculated, using the Pearson Product-Moment Correlation Coefficient (PPMCC), between the class densities and the class accuracies, looking for a correlation between higher accuracy and higher density.  The results of this correlation study are presented in \autoref{tab:density_correlations}, \autoref{tab:density_correlations_mnist}, \autoref{tab:density_correlations_fashion_mnist}, \autoref{tab:density_correlations_imagenette}, and \autoref{tab:density_correlations_cifar10} (a tabulation of the correlation of the individual classes in CIFAR-100 has not been presented here due to the large number of classes).  The study showed that on average, all candidate calculations showed a moderate correlation.  However, the min and max candidates each had a dataset for which the correlation was negative (MNIST and Fashion-MNIST respectively).  Aside from having no datasets with a negative correlation, the mean candidate also had the highest mean correlation.  Notably, CIFAR-10, and CIFAR-100 had the weakest non-negative correlation for all three candidates.  When excluding these datasets, the difference in the mean correlation between the min and max candidates and the mean candidate was even greater.  As such, the mean candidate, \autoref{equation:density_candidate_3}, has been chosen for the calculation of class density.

\begin{table}[!htbp]
\caption{Mean Correlation Between Datasets' Class Accuracies and Class Densities For Each Candidate Class Density Calculation}\label{tab:density_correlations}
\begin{tabularx}{\textwidth}{@{}O{1.7in}XQ{1.3in}Q{1.6in}@{}}
    \toprule
      Density Calculation & Dataset & Correlation & Correlation Excluding CIFAR-10/CIFAR-100 \\
    \midrule
        \multirow{5}{*}{\shortstack{\textit{Min-Derived} Density\\(\autoref{equation:density_candidate_1})}}
        & MNIST         & -0.125230279 & -0.125230279 \\
        & Fashion-MNIST &  0.712199587 &  0.712199587 \\
        & CIFAR-10      &  0.299431773 &          --- \\
        & CIFAR-100     &  0.129662784 &          --- \\
        & Imagenette    &  0.408186019 &  0.408186019 \\
    \midrule
        & Mean          &  0.284849977 &  0.331718442 \\
    \midrule
        \multirow{5}{*}{\shortstack{\textit{Max-Derived} Density\\(\autoref{equation:density_candidate_2})}}
        & MNIST         &  0.627101694 &  0.627101694 \\
        & Fashion-MNIST & -0.105499485 & -0.105499485 \\
        & CIFAR-10      &  0.270797282 &          --- \\
        & CIFAR-100     &  0.153205742 &          --- \\
        & Imagenette    &  0.334691583 &  0.334691583 \\
    \midrule
        & Mean          &  0.256059363 &  0.285431264 \\
    \midrule
        \multirow{5}{*}{\shortstack{\textit{Mean-Derived} Density\\(\autoref{equation:density_candidate_3})}}
        & MNIST         &  0.599024755 &  0.599024755 \\
        & Fashion-MNIST &  0.465716661 &  0.465716661 \\
        & CIFAR-10      &  0.141685897 &          --- \\
        & CIFAR-100     &  0.130301561 &          --- \\
        & Imagenette    &  0.373871535 &  0.373871535 \\
    \midrule
        & Mean & \textbf{0.342120082} & \textbf{0.479537651} \\
    \bottomrule
\end{tabularx}
\end{table}

\begin{table}[!htbp]
    \parbox{.49\linewidth}{
        \centering
        \begin{tabularx}{.49\textwidth}{@{}Crr@{}}
            \toprule
                Class \# & Accuracy & Density (\(d_i\)) \\
            \midrule
                0 & 99.94\% & 1.203606482 \\
                1 & 99.75\% & 0.900390918 \\
                2 & 99.79\% & 0.966957099 \\
                3 & 99.96\% & 1.234064737 \\
                4 & 99.57\% & 0.954383008 \\
                5 & 99.44\% & 0.917783773 \\
                6 & 99.71\% & 1.095102748 \\
                7 & 99.65\% & 1.182528651 \\
                8 & 99.81\% & 0.910899921 \\
                9 & 99.50\% & 0.875168049 \\
            \midrule
                \multicolumn{2}{l}{Correlation Coefficient}
                & 0.599024755 \\
            \bottomrule
        \end{tabularx}
        \caption{Class Accuracy vs. Class Density using the \textit{Mean-Derived} Density Calculation (\autoref{equation:density_candidate_3}) --- MNIST}\label{tab:density_correlations_mnist}
        \vspace{.15in}
    }
    \hfill
    \parbox{.49\linewidth}{
        \centering
        \begin{tabularx}{.49\textwidth}{@{}Crr@{}}
            \toprule
                Class \# & Accuracy & Density (\(d_i\)) \\
            \midrule
                0 & 89.44\% & 0.785806014 \\
                1 & 98.97\% & 0.787244604 \\
                2 & 90.45\% & 0.886734509 \\
                3 & 88.85\% & 0.944081467 \\
                4 & 93.40\% & 0.805763371 \\
                5 & 98.73\% & 0.825402009 \\
                6 & 80.40\% & 0.550174158 \\
                7 & 98.13\% & 1.219426745 \\
                8 & 99.26\% & 0.720677105 \\
                9 & 96.46\% & 1.094532682 \\
            \midrule
                \multicolumn{2}{l}{Correlation Coefficient}
                & 0.465716661 \\
            \bottomrule
        \end{tabularx}
        \caption{Class Accuracy vs. Class Density using the \textit{Mean-Derived} Density Calculation (\autoref{equation:density_candidate_3}) --- Fashion-MNIST}\label{tab:density_correlations_fashion_mnist}
    }
\end{table}

\begin{table}[!htbp]
    \parbox{.49\linewidth}{
        \centering
        \begin{tabularx}{.49\textwidth}{@{}Crr@{}}
            \toprule
                Class \# & Accuracy & Density (\(d_i\)) \\
            \midrule
                0 & 93.68\% & 0.675261028 \\
                1 & 96.49\% & 0.694783936 \\
                2 & 90.34\% & 0.561082899 \\
                3 & 86.04\% & 0.533699944 \\
                4 & 96.20\% & 0.598262085 \\
                5 & 91.48\% & 0.590848128 \\
                6 & 93.88\% & 0.709070212 \\
                7 & 86.70\% & 0.694788326 \\
                8 & 94.08\% & 0.564920444 \\
                9 & 94.95\% & 0.747507803 \\
            \midrule
                \multicolumn{2}{l}{Correlation Coefficient}
                & 0.373871535 \\
            \bottomrule
        \end{tabularx}
        \caption{Class Accuracy vs. Class Density using the \textit{Mean-Derived} Density Calculation (\autoref{equation:density_candidate_3}) --- Imagenette}\label{tab:density_correlations_imagenette}
    }
    \hfill
    \parbox{.49\linewidth}{
        \centering
        \begin{tabularx}{.49\textwidth}{@{}Crr@{}}
            \toprule
                Class \# & Accuracy & Density (\(d_i\)) \\
            \midrule
                0 & 90.97\% & 0.726249735 \\
                1 & 95.01\% & 0.659240091 \\
                2 & 85.27\% & 0.698367900 \\
                3 & 75.59\% & 0.710996520 \\
                4 & 88.89\% & 0.757157015 \\
                5 & 81.51\% & 0.781883942 \\
                6 & 92.85\% & 0.765887353 \\
                7 & 92.78\% & 0.741599817 \\
                8 & 94.28\% & 0.786613742 \\
                9 & 93.99\% & 0.799861845 \\
            \midrule
                \multicolumn{2}{l}{Correlation Coefficient}
                & 0.141685897 \\
            \bottomrule
        \end{tabularx}
        \caption{Class Accuracy vs. Class Density using the \textit{Mean-Derived} Density Calculation (\autoref{equation:density_candidate_3}) --- CIFAR-10}\label{tab:density_correlations_cifar10}
    }
\end{table}

\section{Dynamic Data Reduction}\label{dynamic_data_reduction}

\subsection{Experimental Design}\label{dynamic_data_reduction_experimental_design}

Using the \textit{Mean-Derived} Density Calculation (\autoref{equation:density_candidate_3}), the training data in each class, for each dataset, was reduced using the \textit{Central Exclusion} data reduction strategy from~\cite{Byerly2022} by the number of samples necessary to ensure all classes in the training data that had a density greater than a target density value were reduced by the number of samples needed to achieve that target density value.  In order to find how many samples would need to be excluded, a binary search through each class's samples, ordered by distance from the class's centroid was performed.  The binary search terminated and selected the threshold distance for inclusion after 9 iterations, which was sufficient to choose the number of samples to be included to within a margin of .05\% of the total number of samples in the class.

While studying dynamic data reduction, we used the datasets for which the correlation studies were performed above (MNIST, Fashion-MNIST, CIFAR-10, CIFAR-100, and Imagenette) as well as two additional datasets.  The first is the micro-PCB dataset introduced in~\cite{Byerly2021c}.  Data augmentation used during the training for the micro-PCB dataset was uniform and the same strategy as used for the experiments in~\cite{Byerly2021a}.  The second additional dataset was EMNIST-Digits~\cite{Cohen2017}, which is a dataset with the same format as MNIST and with the same domain of interest, namely the Hindu-Arabic numerals.  The difference is that EMNIST-Digits contains exactly four times as many samples in both the training and test sets.  This allows for an investigation into whether the test accuracy achieved after reducing data to a target threshold per class is sensitive to or independent of the number of samples used for training.  Data augmentation used during the training for EMNIST-Digits was uniform and the same strategy as used for the experiments in~\cite{Byerly2021b}.

Each of MNIST, Fashion-MNIST, CIFAR-10, CIFAR-100, and EMNIST-Digits were subjected to this method using target density levels of 1.0, 0.9, 0.8, 0.7, 0.6, and 0.5.  These target densities were chosen because they produced a similar number of excluded samples for these datasets.  However, in the case of Imagenette and the micro-PCB dataset, the UMAP dimensional reduction produced values in each dimension with much larger variances (thus producing smaller densities).  This is due to the fact that the UMAP algorithm assumes that the population in the higher dimensional space from which the data is sampled is uniformly distributed on some manifold.  Since Imagenette and the micro-PCB dataset are both higher dimensional than the other datasets as well as having fewer samples in the training set, UMAP assumes the values of these samples in the higher dimensional space represent a much broader range of possible values for the assumed uniformity of the population.  Therefore, a larger number of approximately evenly spaced target densities was chosen specifically for each of Imagenette and of the micro-PCB dataset.

For all experiments, 5 trials were conducted, and the accuracy reported is the mean accuracy of those 5 trials.

\subsection{Experimental Results}\label{dynamic_data_reduction_experimental_results}

\autoref{tab:dynamic_density1_MNIST} through \autoref{tab:dynamic_density1_EMNISTDigits} show the results of these experiments.  Bold typeface in the Accuracy columns indicate that the accuracy matched or exceeded the accuracy of the baseline.

For the MNIST dataset, there was no statistically significant difference (\(p < 0.05\)) between the accuracy of the baseline and the accuracies of the experiments with target densities 1.0, 0.9, 0.8, and 0.7.  These target densities resulted in reducing the training dataset size by 4.4\%, 7.8\%, 15\%, and 22.6\%, respectively.

For the Fashion-MNIST dataset, there was no statistically significant difference (\(p < 0.05\)) between the accuracy of the baseline and the accuracy of the experiment with target density 1.0 which resulted in reducing the training dataset size by 2.0\%.

For the CIFAR-10 dataset, there was no statistically significant difference (\(p < 0.05\)) between the accuracy of the baseline and the accuracies of the experiments with target densities 1.0, 0.9, 0.8, and 0.7.  These target densities resulted in reducing the training dataset size by 0.2\%, 0.2\%, 0.2\%, and 4.2\%, respectively.

For the CIFAR-100 dataset, there was no statistically significant difference (\(p < 0.05\)) between the accuracy of the baseline and the accuracies of the experiments with target densities 1.0 and 0.8.  These target densities resulted in reducing the training dataset size by 0.4\% and 2.3\%, respectively,  The experiment with a target density of 0.9 produced a statistically significantly superior accuracy while reducing the dataset size by 0.4\%.

For the Imagenette dataset, there was no statistically significant difference (\(p < 0.05\)) between the accuracy of the baseline and the accuracies of the experiments with target densities 0.70, 0.65, 0.60, and 0.55.  These target densities resulted in reducing the training dataset size by 0.3\%, 2.4\%, 5.1\%, and 9.2\%, respectively.

For the micro-PCB dataset, there was no statically significant difference (\(p < 0.05\)) between the accuracy of the baseline and the accuracies of the experiments with target densities 0.25, 0.20, 0.15, 0.10, 0.075, and 0.05.  These target densities resulted in reducing the training dataset size by 5.4\%, 15.8\%, 30.5\%, 42.9\%, and 53.2\%, respectively.  Of particular note with the micro-PCB dataset is that the baseline accuracy of 100\% was able to be achieved with 53.2\% of the training samples removed.

For the EMNIST-Digits dataset, there was no statistically significant difference (\(p < 0.05\)) between the accuracy of the baseline and the accuracies of the experiments with target densities 1.0, 0.9, and 0.8.  These target densities resulted in reducing the training dataset size by 3.1\%, 7.2\%, and 13.4\%, respectively.

\begin{table}[!htbp]
  \caption{Class Accuracies Achieved at Target Densities --- MNIST}\label{tab:dynamic_density1_MNIST}
  \begin{tabularx}{\textwidth}{@{}cCrrr@{}}
  \toprule
      Target Density (\(d_i\)) & \# Samples Included & Accuracy & Std. Dev. & p-value \\
  \midrule
      N/A   & 100.0\% &           99.716\% & 0.000162481 & ---         \\
      1.0   &  95.6\% &           99.714\% & 0.000080000 & 0.415365855 \\
      0.9   &  92.2\% &  \textbf{99.732\%} & 0.000172047 & 0.106638807 \\
      0.8   &  85.0\% &           99.708\% & 0.000097980 & 0.21178501  \\
      0.7   &  77.4\% &  \textbf{99.716\%} & 0.000135647 & 0.5         \\
      0.6   &  69.4\% &           99.682\% & 0.000116619 & 0.004680234 \\
      0.5   &  61.0\% &           99.694\% & 0.000080000 & 0.020617404 \\
  \bottomrule
  \end{tabularx}
\end{table}

\begin{table}[!htbp]
  \caption{Class Accuracies Achieved at Target Densities --- Fashion-MNIST}\label{tab:dynamic_density1_FashionMNIST}
  \begin{tabularx}{\textwidth}{@{}cCrrr@{}}
  \toprule
    Target Density (\(d_i\)) & \# Samples Included & Accuracy & Std. Dev. & p-value \\
  \midrule
    N/A   & 100.0\% &  93.404\% & 0.001380724 & ---           \\
    1.0   &  98.0\% &  93.298\% & 0.000928224 & 0.119171827   \\
    0.9   &  96.4\% &  93.202\% & 0.000982649 & 0.022137464   \\
    0.8   &  93.2\% &  93.254\% & 0.000611882 & 0.041111842   \\
    0.7   &  86.4\% &  92.994\% & 0.001330564 & 0.001350314   \\
    0.6   &  78.2\% &  92.574\% & 0.001089220 & \num{6.52e-6} \\
    0.5   &  68.8\% &  91.922\% & 0.001750885 & \num{4.90e-7} \\
  \bottomrule
  \end{tabularx}
\end{table}

\begin{table}[!htbp]
  \caption{Class Accuracies Achieved at Target Densities --- CIFAR-10}\label{tab:dynamic_density1_Cifar10}
  \begin{tabularx}{\textwidth}{@{}cCrrr@{}}
  \toprule
    Target Density (\(d_i\)) & \# Samples Included & Accuracy & Std. Dev. & p-value \\
  \midrule
    N/A   & 100.0\% &           89.146\% & 0.001518684 & ---           \\
    1.0   &  99.8\% &  \textbf{89.342\%} & 0.002688048 & 0.119942536   \\
    0.9   &  99.8\% &  \textbf{89.154\%} & 0.002361864 & 0.477979183   \\
    0.8   &  99.8\% &  \textbf{89.346\%} & 0.002151836 & 0.083656408   \\
    0.7   &  95.8\% &  \textbf{89.192\%} & 0.001828004 & 0.354382759   \\
    0.6   &  86.2\% &           88.700\% & 0.002044505 & 0.004518233   \\
    0.5   &  75.4\% &           88.064\% & 0.002030369 & \num{1.37e-5} \\
  \bottomrule
  \end{tabularx}
\end{table}

\begin{table}[!htbp]
  \caption{Class Accuracies Achieved at Target Densities --- CIFAR-100}\label{tab:dynamic_density1_Cifar100}
  \begin{tabularx}{\textwidth}{@{}cCrrr@{}}
  \toprule
    Target Density (\(d_i\)) & \# Samples Included & Accuracy & Std. Dev. & p-value \\
  \midrule
    N/A   & 100.0\% &           61.896\% & 0.001786169 & ---           \\
    1.0   &  99.6\% &  \textbf{62.186\%} & 0.003273286 & 0.079228788   \\
    0.9   &  99.6\% &  \textbf{62.220\%} & 0.002830548 & 0.044442483   \\
    0.8   &  97.7\% &           61.876\% & 0.003501200 & 0.460722547   \\
    0.7   &  93.1\% &           61.642\% & 0.002057571 & 0.049626893   \\
    0.6   &  85.4\% &           60.398\% & 0.003592993 & \num{3.58e-5} \\
    0.5   &  75.1\% &           59.108\% & 0.001561282 & \num{5.71e-9} \\
  \bottomrule
  \end{tabularx}
\end{table}

\begin{table}[!htbp]
  \caption{Class Accuracies Achieved at Target Densities --- Imagenette}\label{tab:dynamic_density1_imagenette}
  \begin{tabularx}{\textwidth}{@{}cCrrr@{}}
  \toprule
    Target Density (\(d_i\)) & \# Samples Included & Accuracy & Std. Dev. & p-value \\
  \midrule
    N/A   & 100.0\% &           92.390\% & 0.002333238 & ---            \\
    0.70  &  99.7\% &           92.104\% & 0.002514438 & 0.06698284     \\
    0.65  &  97.6\% &  \textbf{92.486\%} & 0.001497465 & 0.25410172     \\
    0.60  &  94.9\% &           92.250\% & 0.003331066 & 0.255309713    \\
    0.55  &  90.8\% &           92.120\% & 0.002728369 & 0.085472681    \\
    0.50  &  85.1\% &           91.798\% & 0.001984339 & 0.002385587    \\
    0.45  &  78.6\% &           91.434\% & 0.001416474 & \num{5.60e-5}  \\
    0.40  &  72.2\% &           90.650\% & 0.002086145 & \num{1.91e-6}  \\
    0.35  &  65.1\% &           90.026\% & 0.003338622 & \num{1.38e-6}  \\
    0.30  &  57.6\% &           89.626\% & 0.001276871 & \num{1.51e-8}  \\
    0.25  &  49.8\% &           88.206\% & 0.002298347 & \num{2.95e-9}  \\
    0.20  &  41.7\% &           86.304\% & 0.005313229 & \num{1.40e-8}  \\
    0.15  &  32.5\% &           83.866\% & 0.00475546  & \num{4.73e-10} \\
    0.10  &  22.4\% &           79.134\% & 0.004674441 & \num{1.26e-11} \\
    0.05  &  9.15\% &           43.564\% & 0.037764248 & \num{2.72e-9}  \\
  \bottomrule
  \end{tabularx}
\end{table}

\begin{table}[!htbp]
  \caption{Class Accuracies Achieved at Target Densities --- micro-PCB}\label{tab:dynamic_density1_microPCB}
  \begin{tabularx}{\textwidth}{@{}cCrrr@{}}
  \toprule
    Target Density (\(d_i\)) & \# Samples Included & Accuracy & Std. Dev. & p-value \\
  \midrule
    N/A   &         100.0\% &          100.000\% &              0 & ---            \\
    0.25  &          94.6\% & \textbf{100.000\%} &              0 & N/A            \\
    0.20  &          84.2\% & \textbf{100.000\%} &              0 & N/A            \\
    0.15  &          69.5\% & \textbf{100.000\%} &              0 & N/A            \\
    0.10  &          57.1\% & \textbf{100.000\%} &              0 & N/A            \\
    0.075 & \textbf{46.8\%} & \textbf{100.000\%} &              0 & N/A            \\
    0.05  &          34.5\% &           99.912\% & \num{1.48e-03} & 0.133988429    \\
    0.04  &          25.6\% &           99.024\% & \num{9.49e-03} & 0.036866271    \\
    0.03  &          17.2\% &           75.112\% & \num{5.52e-02} & \num{9.14e-6}  \\
    0.02  &          11.8\% &           48.088\% & \num{2.24e-01} & 0.000846406    \\
    0.01  &          6.90\% &           18.800\% & \num{5.45e-02} & \num{8.78e-10} \\
  \bottomrule
  \end{tabularx}
\end{table}

\begin{table}[!htbp]
  \caption{Class Accuracies Achieved at Target Densities --- EMNIST-Digits}\label{tab:dynamic_density1_EMNISTDigits}
  \begin{tabularx}{\textwidth}{@{}cCrrr@{}}
  \toprule
    Target Density (\(d_i\)) & \# Samples Included & Accuracy & Std. Dev. & p-value \\
  \midrule
    N/A   & 100.0\% &           99.790\% & 0           & ---           \\
    1.0   &  96.9\% &  \textbf{99.790\%} & 6.32456E-05 & 0.5           \\
    0.9   &  92.8\% &           99.786\% & 4.89898E-05 & 0.070556641   \\
    0.8   &  86.6\% &           99.782\% & 9.79796E-05 & 0.070556641   \\
    0.7   &  79.8\% &           99.784\% & 4.89898E-05 & 0.019984262   \\
    0.6   &  71.5\% &           99.772\% & 4.00000E-05 & \num{9.27e-6} \\
    0.5   &  62.8\% &           99.766\% & 1.01980E-04 & 0.000764001   \\
  \bottomrule
  \end{tabularx}
\end{table}

\section{Dynamic Data Reduction Redux}\label{dynamic_data_reduction_redux}

\subsection{Experimental Design}\label{dynamic_data_reduction_redux_experimental_design}

As was noted in the previous section, the UMAP dimensionality reduction algorithm makes assumptions about the distribution of the data in the higher dimensional space that results in density values that are dependent on the dataset.  In order to account for this, the \textit{Mean-Derived} Density Calculation (\autoref{equation:density_candidate_3}) has been modified to include a normalization term.  Additionally, the first three terms of \autoref{equation:density_candidate_3} have been elided as the purpose they served of biasing unbalanced classes does not apply to the well-balanced datasets being studied here.

In \autoref{equation:density_candidate_4}, \(\overline{d_i}\) is the density of class \(i\) among all \(n\) classes.  \(\sigma_i\) and \(\sigma_j\) are the standard deviations of the \(m\) gaussians of the \(m\)-dimensional reduction for classes \(i\) and \(j\), respectively.

\begin{figure}[!htbp]
  \centering
  \begin{equation}
    \overline{d_i} = 
          \frac{1}{n}
          \sum^n_{j}
          {\left(\frac{1}{m}\sum^m_{k}\sigma_{j_{k}}\right)}
          \cdot {\left(\frac{1}{m}\sum^m_{k}\sigma_{i_{k}}\right)}^{-1}
          \label{equation:density_candidate_4}
  \end{equation}
  \caption{\textit{Mean-Derived} and Normalized Density Calculation}
\end{figure}

\autoref{tab:density_calc_comparison_mnist} through \autoref{tab:density_calc_comparison_fashion_mnist} show a comparison of the densities calculated using the \textit{Mean-Derived} Density Calculation (\autoref{equation:density_candidate_3}) vs.\ the
\textit{Mean-Derived} and Normalized Density Calculation (\autoref{equation:density_candidate_4}) (a comparison of the individual classes in CIFAR-100 has not been tabulated here due to the large number of classes).

As in the previous section, but instead using the \textit{Mean-Derived} and Normalized Density Calculation (\autoref{equation:density_candidate_4}), another set of experiments was conducted.  In these experiments, the training data in each class was reduced using the \textit{Central Exclusion} data reduction strategy by the number of samples necessary to ensure all classes in the training data that had a density greater than a target density value were reduced by the number of samples needed to achieve that target density value.  Each of MNIST, Fashion-MNIST, CIFAR-10, CIFAR-100, Imagenette, the micro-PCB dataset, and EMNIST-Digits were subjected to this method using target density levels of 1.1, 1.05, 1.0, 0.95, and 0.9.

For all experiments, 5 trials were conducted, and the accuracy reported is the mean accuracy of those 5 trials.

\begin{table}[!htbp]
  \parbox{.49\linewidth}{
      \centering
      \begin{tabularx}{.49\textwidth}{@{}Xrr@{}}
          \toprule
              \small{\shortstack{\\Class \#}} & \small{\shortstack{\\Mean-Derived}} & \small{\shortstack{Mean-Derived\\and Normalized}} \\
          \midrule
              0 & 1.203606 & 1.208877 \\
              1 & 0.900391 & 0.794478 \\
              2 & 0.966957 & 0.965486 \\
              3 & 1.234065 & 1.197419 \\
              4 & 0.954383 & 0.971853 \\
              5 & 0.917784 & 1.007165 \\
              6 & 1.095103 & 1.100828 \\
              7 & 1.182529 & 1.122872 \\
              8 & 0.910900 & 0.926147 \\
              9 & 0.875168 & 0.875159 \\
          \bottomrule
      \end{tabularx}
      \caption{\textit{Mean-Derived} Density Calculation (\autoref{equation:density_candidate_3}) vs. \textit{Mean-Derived} and Normalized Density Calculation (\autoref{equation:density_candidate_4}) --- MNIST}\label{tab:density_calc_comparison_mnist}
  }
  \hfill
  \parbox{.49\linewidth}{
    \centering
    \begin{tabularx}{.49\textwidth}{@{}Xrr@{}}
        \toprule
            \small{\shortstack{\\Class \#}} & \small{\shortstack{\\Mean-Derived}} & \small{\shortstack{Mean-Derived\\and Normalized}} \\
        \midrule
            0 & 1.184466 & 1.225866 \\
            1 & 0.696331 & 0.720669 \\
            2 & 1.045727 & 1.082278 \\
            3 & 1.163892 & 1.204573 \\
            4 & 0.963187 & 0.996853 \\
            5 & 0.963509 & 0.997186 \\
            6 & 1.118438 & 1.157530 \\
            7 & 0.996212 & 1.031032 \\
            8 & 0.904710 & 0.936332 \\
            9 & 0.851514 & 0.881277 \\
        \bottomrule
    \end{tabularx}
    \caption{\textit{Mean-Derived} Density Calculation (\autoref{equation:density_candidate_3}) vs. \textit{Mean-Derived} and Normalized Density Calculation (\autoref{equation:density_candidate_4}) --- EMNIST-Digits}\label{tab:density_calc_comparison_emnist_digits}
}
\end{table}

\begin{table}[!htbp]
  \parbox{.49\linewidth}{
      \centering
      \begin{tabularx}{.49\textwidth}{@{}Xrr@{}}
          \toprule
              \small{\shortstack{\\Class \#}} & \small{\shortstack{\\Mean-Derived}} & \small{\shortstack{Mean-Derived\\and Normalized}} \\
          \midrule
              0 & 0.675261 & 1.054048 \\
              1 & 0.694784 & 1.093608 \\
              2 & 0.561083 & 0.849362 \\
              3 & 0.533700 & 0.935029 \\
              4 & 0.598262 & 0.955690 \\
              5 & 0.590848 & 0.929037 \\
              6 & 0.709070 & 1.109126 \\
              7 & 0.694788 & 1.121807 \\
              8 & 0.564920 & 0.892939 \\
              9 & 0.747508 & 1.170468 \\
          \bottomrule
      \end{tabularx}
      \caption{\textit{Mean-Derived} Density Calculation (\autoref{equation:density_candidate_3}) vs. \textit{Mean-Derived} and Normalized Density Calculation (\autoref{equation:density_candidate_4}) --- Imagenette}\label{tab:density_calc_comparison_imagenette}
  }
  \hfill
  \parbox{.49\linewidth}{
    \centering
    \begin{tabularx}{.49\textwidth}{@{}Xrr@{}}
        \toprule
            \small{\shortstack{\\Class \#}} & \small{\shortstack{\\Mean-Derived}} & \small{\shortstack{Mean-Derived\\and Normalized}} \\
        \midrule
            0 & 0.726250 & 0.980983 \\
            1 & 0.659240 & 0.890469 \\
            2 & 0.698368 & 0.943321 \\
            3 & 0.710997 & 0.960379 \\
            4 & 0.757157 & 1.022731 \\
            5 & 0.781884 & 1.056131 \\
            6 & 0.765887 & 1.034523 \\
            7 & 0.741600 & 1.001717 \\
            8 & 0.786614 & 1.062520 \\
            9 & 0.799862 & 1.080414 \\
        \bottomrule
    \end{tabularx}
    \caption{\textit{Mean-Derived} Density Calculation (\autoref{equation:density_candidate_3}) vs. \textit{Mean-Derived} and Normalized Density Calculation (\autoref{equation:density_candidate_4}) --- CIFAR-10}\label{tab:density_calc_comparison_cifar10}
}
\end{table}

\begin{table}[!htbp]
  \begin{minipage}[t]{\linewidth}
    \parbox{.49\linewidth}{
        \centering
        \begin{tabularx}{.49\textwidth}{@{}Xrr@{}}
            \toprule
                \small{\shortstack{\\Class \#}} & \small{\shortstack{\\Mean-Derived}} & \small{\shortstack{Mean-Derived\\and Normalized}} \\
            \midrule
                0 & 0.277563 & 1.007280 \\
                1 & 0.260718 & 0.946147 \\
                2 & 0.255069 & 0.925648 \\
                3 & 0.264466 & 0.959749 \\
                4 & 0.273648 & 0.993069 \\
                5 & 0.308335 & 1.118948 \\
                6 & 0.274374 & 0.995703 \\
                7 & 0.286130 & 1.038367 \\
                8 & 0.281862 & 1.022877 \\
                9 & 0.278199 & 1.009587 \\
               10 & 0.276846 & 1.004674 \\
               11 & 0.276255 & 1.002530 \\
               12 & 0.275822 & 1.000960 \\
            \bottomrule
        \end{tabularx}
        \caption{\textit{Mean-Derived} Density Calculation (\autoref{equation:density_candidate_3}) vs. \textit{Mean-Derived} and Normalized Density Calculation (\autoref{equation:density_candidate_4}) --- micro-PCB}\label{tab:density_calc_comparison_micro_pcb}
    }
    \hfill
    \parbox{.49\linewidth}{
      \centering
      \begin{tabularx}{.49\textwidth}{@{}Xrr@{}}
          \toprule
              \small{\shortstack{\\Class \#}} & \small{\shortstack{\\Mean-Derived}} & \small{\shortstack{Mean-Derived\\and Normalized}} \\
          \midrule
              0 & 0.785806 & 0.952496 \\
              1 & 0.787245 & 0.954240 \\
              2 & 0.886735 & 1.074834 \\
              3 & 0.944081 & 1.144346 \\
              4 & 0.805763 & 0.976687 \\
              5 & 0.825402 & 1.000492 \\
              6 & 0.550174 & 0.666881 \\
              7 & 1.219427 & 1.478100 \\
              8 & 0.720677 & 0.873552 \\
              9 & 1.094533 & 1.326712 \\
          \bottomrule
      \end{tabularx}
      \vspace{.45in}
      \caption{\textit{Mean-Derived} Density Calculation (\autoref{equation:density_candidate_3}) vs. \textit{Mean-Derived} and Normalized Density Calculation (\autoref{equation:density_candidate_4}) --- Fashion-MNIST}\label{tab:density_calc_comparison_fashion_mnist}
    }
  \end{minipage}
\end{table}

\subsection{Experimental Results}\label{dynamic_data_reduction_redux_experimental_results}

\autoref{tab:dynamic_density2_MNIST} through \autoref{tab:dynamic_density2_EMNISTDigits} show the results of these experiments.  Bold typeface in the Accuracy columns indicate that the accuracy matched or exceeded the accuracy of the baseline.

For the MNIST dataset, there was no statistically significant difference (\(p < 0.05\)) between the accuracy of the baseline and the accuracies of any of the experiments with target densities.  These target densities resulted in reducing the training dataset size by 4.2\%, 8.0\%, 11.8\%, 17.2\%, and 24.2\%, respectively.

For the Fashion-MNIST dataset, all target densities were statistically significantly inferior to the baseline (\(p < 0.05\)).  This is consistent with the results from the prior section as the highest target density resulted in removing 15\% of the data whereas the only statistically insignificant data reduction using \autoref{equation:density_candidate_3} elided just 2\% of the data.

For the CIFAR-10 dataset, there was no statistically significant difference (\(p < 0.05\)) between the accuracy of the baseline and the accuracies of the experiments with target densities 1.1, 1.05, and 1.0.  These target densities resulted in reducing the training dataset size by 0.2\%, 1.1\%, and 5.4\%, respectively.

For the CIFAR-100 dataset, all target densities were statistically significantly inferior to the baseline (\(p < 0.05\)).  This is consistent with the results from the prior section as the highest target density resulted in removing 5.7\% of the data whereas statistically insignificant data reduction using \autoref{equation:density_candidate_3} was only achieved when eliding 2.3\% of the data.

For the Imagenette dataset, there was no statistically significant difference (\(p < 0.05\)) between the accuracy of the baseline and the accuracies of the experiments with target density 1.1.  This target density resulted in reducing the training dataset size by 1.7\%.  This is inconsistent with the results from the prior section as statistically insignificant data reduction using \autoref{equation:density_candidate_3} was achieved when eliding as much as 9.2\% of the data.

For the micro-PCB dataset, all target densities achieved 100\% test accuracy while reducing the training dataset size by 0.5\%, 0.5\%, 2\%, 8.4\%, and 19.9\%, respectively.

For the EMNIST-Digits dataset, there was no statistically significant difference (\(p < 0.05\)) between the accuracy of the baseline and the accuracies of the experiments with target densities 1.1 and 1.05.  These target densities resulted in reducing the training dataset size by 5.1\% and 8.5\%, respectively.  After 5 trials, a statistical significance to the superior accuracy of the experiment with target density of 1.0 was narrowly missed (\(p < 0.07\)).  Of worthy note for this dataset is the remarkable consistency and low variance across trials for the baseline (which all achieved the same accuracy of 99.79\%) and across trials for each of the target densities.  The trials for target density 1.1 produced two accuracies of 99.79\% and three accuracies of 99.78\%.  The trials for target density 1.05 produced one accuracy of 99.79\%, three accuracies of 99.78\%, and the lowest accuracy of all trials of 99.76\%.  The trials for target density 1.0 produced two of the highest accuracies at 99.8\% and three at 99.79\%.  The trials for target density 0.95 produced three accuracies of 99.78\% and two of 99.77\%.  The trials for target density 0.9 produced one accuracy of 99.79\%, two accuracies of 99.78\%, and two accuracies of 99.77\%.

\begin{table}[!htbp]
  \caption{Class Accuracies Achieved at Target Densities --- MNIST}\label{tab:dynamic_density2_MNIST}
  \begin{tabularx}{\textwidth}{@{}cCrrr@{}}
  \toprule
      Target Density (\(\overline{d_i}\)) & \# Samples Included & Accuracy & Std. Dev. & p-value \\
  \midrule
      N/A   & 100.0\% &         99.716\%  & 0.000162481 & ---         \\
      1.10  &  95.8\% & \textbf{99.722\%} & 0.000172047 & 0.312884652 \\
      1.05  &  92.0\% &         99.714\%  & 0.000101980 & 0.420019164 \\
      1.00  &  88.2\% & \textbf{99.730\%} & 0.000209762 & 0.161058701 \\
      0.95  &  82.8\% & \textbf{99.720\%} & 0.000167332 & 0.370219727 \\
      0.90  &  75.8\% &         99.706\%  & 0.000185472 & 0.220382883 \\
  \bottomrule
  \end{tabularx}
\end{table}

\begin{table}[!htbp]
  \caption{Class Accuracies Achieved at Target Densities --- Fashion-MNIST}\label{tab:dynamic_density2_FashionMNIST}
  \begin{tabularx}{\textwidth}{@{}cCrrr@{}}
  \toprule
    Target Density (\(\overline{d_i}\)) & \# Samples Included & Accuracy & Std. Dev. & p-value \\
  \midrule
    N/A   & 100.0\% &  93.404\% & 0.001380724 & ---            \\
    1.10  &  85.0\% &  84.058\% & 0.004762100 & \num{1.35e-10} \\
    1.05  &  82.6\% &  83.752\% & 0.001151347 & \num{3.16e-14} \\
    1.00  &  74.6\% &  83.526\% & 0.001400857 & \num{5.39e-14} \\
    0.95  &  74.2\% &  82.834\% & 0.001293986 & \num{2.30e-14} \\
    0.90  &  65.8\% &  81.716\% & 0.002239286 & \num{1.44e-13} \\
  \bottomrule
  \end{tabularx}
\end{table}

\begin{table}[!htbp]
  \caption{Class Accuracies Achieved at Target Densities --- CIFAR-10}\label{tab:dynamic_density2_Cifar10}
  \begin{tabularx}{\textwidth}{@{}cCrrr@{}}
  \toprule
    Target Density (\(\overline{d_i}\)) & \# Samples Included & Accuracy & Std. Dev. & p-value \\
  \midrule
    N/A   & 100.0\% &         89.146\%  & 0.001518684 & ---            \\
    1.10  &  99.8\% &         89.104\%  & 0.001504128 &    0.352296774 \\
    1.05  &  98.9\% & \textbf{89.166\%} & 0.002129413 &    0.441118341 \\
    1.00  &  94.6\% &         89.130\%  & 0.002399167 &    0.456523145 \\
    0.95  &  86.6\% &         88.742\%  & 0.002066301 &    0.006790324 \\
    0.90  &  75.8\% &         88.140\%  & 0.001052616 & \num{2.24e-06} \\
  \bottomrule
  \end{tabularx}
\end{table}

\begin{table}[!htbp]
  \caption{Class Accuracies Achieved at Target Densities --- CIFAR-100}\label{tab:dynamic_density2_Cifar100}
  \begin{tabularx}{\textwidth}{@{}cCrrr@{}}
  \toprule
    Target Density (\(\overline{d_i}\)) & \# Samples Included & Accuracy & Std. Dev. & p-value \\
  \midrule
    N/A   & 100.0\% & 61.896\% & 0.001786169 & ---            \\
    1.10  &  94.3\% & 61.194\% & 0.002620382 &    0.00110249  \\
    1.05  &  90.2\% & 60.736\% & 0.001651181 & \num{6.04e-06} \\
    1.00  &  84.5\% & 58.470\% & 0.001255388 & \num{5.78e-10} \\
    0.95  &  78.0\% & 57.134\% & 0.004187410 & \num{1.43e-08} \\
    0.90  &  70.1\% & 55.368\% & 0.003521023 & \num{3.81e-10} \\
  \bottomrule
  \end{tabularx}
\end{table}

\begin{table}[!htbp]
  \caption{Class Accuracies Achieved at Target Densities --- Imagenette}\label{tab:dynamic_density2_imagenette}
  \begin{tabularx}{\textwidth}{@{}cCrrr@{}}
  \toprule
    Target Density (\(\overline{d_i}\)) & \# Samples Included & Accuracy & Std. Dev. & p-value \\
  \midrule
    N/A   & 100.0\% & 92.390\% & 0.002333238 & ---            \\
    1.10  &  98.3\% & 92.224\% & 0.000705975 & 0.105163732    \\
    1.05  &  94.2\% & 92.126\% & 0.001473228 & 0.04602054     \\
    1.00  &  88.8\% & 92.080\% & 0.000748331 & 0.017619593    \\
    0.95  &  82.7\% & 91.316\% & 0.002465441 & 0.000112946    \\
    0.90  &  73.9\% & 90.428\% & 0.002318103 & \num{1.12e-06} \\
  \bottomrule
  \end{tabularx}
\end{table}

\begin{table}[!htbp]
  \caption{Class Accuracies Achieved at Target Densities --- micro-PCB}\label{tab:dynamic_density2_microPCB}
  \begin{tabularx}{\textwidth}{@{}cCrrr@{}}
  \toprule
    Target Density (\(\overline{d_i}\)) & \# Samples Included & Accuracy & Std. Dev. & p-value \\
  \midrule
    N/A   & 100.0\% &         100.000\%  & 0 & --- \\
    1.10  &  99.5\% & \textbf{100.000\%} & 0 & N/A \\
    1.05  &  99.5\% & \textbf{100.000\%} & 0 & N/A \\
    1.00  &  98.0\% & \textbf{100.000\%} & 0 & N/A \\
    0.95  &  91.6\% & \textbf{100.000\%} & 0 & N/A \\
    0.90  &  80.1\% & \textbf{100.000\%} & 0 & N/A \\
  \bottomrule
  \end{tabularx}
\end{table}

\begin{table}[!htbp]
  \caption{Class Accuracies Achieved at Target Densities --- EMNIST-Digits}\label{tab:dynamic_density2_EMNISTDigits}
  \begin{tabularx}{\textwidth}{@{}cCrrr@{}}
  \toprule
    Target Density (\(\overline{d_i}\)) & \# Samples Included & Accuracy & Std. Dev. & p-value \\
  \midrule
    N/A   & 100.0\% &          99.790\% & 0              & ---         \\
    1.10  &  94.9\% &          99.784\% & \num{4.90e-05} & 0.019984262 \\
    1.05  &  91.5\% &          99.778\% & \num{9.80e-05} & 0.019984262 \\
    1.00  &  87.2\% & \textbf{99.794\%} & \num{4.90e-05} & 0.070556641 \\
    0.95  &  80.7\% &          99.776\% & \num{4.90e-05} & 0.000223176 \\
    0.90  &  73.4\% &          99.778\% & \num{7.48e-05} & 0.006238937 \\
  \bottomrule
  \end{tabularx}
\end{table}

\section{Dataset Quality}\label{dataset_quality}

The experiments detailed thus far have demonstrated that, for the datasets studied, there exists an amount of redundant data in each such dataset and that that data can be located near the centroid of a dimensional reduction of the data.

This observation inspires the possibility of a measure of the dataset quality that can be used \textit{a-priori} to determine whether or not the data distributed in the lower-dimensional space is capable of leading to optimal or near-optimal classification accuracy.

\autoref{equation:dataset_quality} represents an attempt to capture this measure numerically in a single value, such that the closer the value is to zero, the lower the ``quality'' of the dataset, or from another point of view, there exists room for improvement.  In this equation, \(d\) are the densities of the classes as calculated using \autoref{equation:density_candidate_4}.  As such, the equation is simply 1 over the product of the standard deviation of the class densities and the range of density values.  Thus and simply, the greater variance among the class densities, the lower the quality of the data.

\begin{figure}[!htbp]
  \centering
  \begin{equation}
      q = \frac{1}{\sigma_{d}\cdot\left(\max\left(d\right)-\min\left(d\right)\right)}
      \label{equation:dataset_quality}
    \end{equation}
  \caption{Dataset Quality Calculation}
\end{figure}

\autoref{tab:train_dataset_quality_comparison} shows this calculation as applied to the training datasets we have studied.  MNIST, EMNIST-Digits, CIFAR-10 and the micro-PCB dataset all had values for quality \(> 10\) and all of these datasets were shown to have had excess data in their training datasets.  Fashion-MNIST and CIFAR-100 each had values for quality \(< 10\) and both of these datasets were shown not to have had an excess of data in their training datasets.  Imagenette was inconsistent with this.  It is hypothesized that in the case of Imagenette, that the lower number of training samples coupled with the complexity of the features that comprise the subject matter, compared to the other datasets, dominates any ability to elide redundant data.  \autoref{tab:test_dataset_quality_comparison} shows the same calculation as applied to the validation data of the studied datasets.  Consistent with the training data, the validation data had values for quality \(> 10\) for MNIST, EMNIST-Digits, CIFAR-10 and the micro-PCB dataset, and Fashion-MNIST and CIFAR-100 each had values for quality \(< 10\).


\begin{table}[!htbp]
  \caption{Statistical Properties and the Quality of Training Datasets}\label{tab:train_dataset_quality_comparison}
  \begin{tabularx}{\textwidth}{@{}Xrrr@{}}
      \toprule
        Training Dataset & Std. Dev. & Range & Quality \\
      \midrule
        MNIST         & 0.1304920 & 0.4143995 &  18.492560 \\
        EMNIST-Digits & 0.1470203 & 0.5051971 &  13.463625 \\
        Fashion-MNIST & 0.2176022 & 0.8112188 &   5.664984 \\
        CIFAR-10      & 0.0568050 & 0.1899450 &  92.680124 \\
        CIFAR-100     & 0.1623155 & 0.8421434 &   7.315669 \\
        Imagenette    & 0.1056541 & 0.3211061 &  29.475744 \\
        micro-PCB     & 0.0450495 & 0.1933003 & 114.835613 \\
      \bottomrule
  \end{tabularx}
\end{table}

\begin{table}[!htbp]
  \caption{Statistical Properties and the Quality of Validation Datasets}\label{tab:test_dataset_quality_comparison}
  \begin{tabularx}{\textwidth}{@{}Xrrr@{}}
      \toprule
        Validation Dataset & Std. Dev. & Range & Quality \\
      \midrule
        MNIST         & 0.1308110 & 0.4797679 &  15.933995 \\
        EMNIST-Digits & 0.1268322 & 0.3672039 &  21.471530 \\
        Fashion-MNIST & 0.2274090 & 0.8411320 &   5.227911 \\
        CIFAR-10      & 0.0495897 & 0.1651205 & 122.125472 \\
        CIFAR-100     & 0.1321202 & 0.7995342 &   9.466595 \\
        Imagenette    & 0.1223012 & 0.3698493 &  22.107753 \\
        micro-PCB     & 0.1455047 & 0.4808525 & 14.2925951 \\
      \bottomrule
  \end{tabularx}
\end{table}

\section{Summary}

In this work, we have put forth a definition for class density along with several methods for calculating such.  Analysis of seven datasets showed that \autoref{equation:density_candidate_4}, derived from the normalized mean standard deviations of the data points in lower dimensional space, was the appropriate choice for the calculation for density.  Additionally, we put forth a definition for dataset quality in \autoref{equation:dataset_quality}, which showed that those datasets that met a certain quality threshold (experimentally demonstrated to be \(> 10\) for the datasets studied) were candidates for eliding redundant data using \autoref{equation:density_candidate_4}.  The experiments showed that for all datasets except Imagenette, datasets with a quality of \(> 10\) could be reduced to a target density of at most 1.0 and achieve accuracy that is statistically insignificantly different from the baseline.  Unique to Imagenette is the relatively low number of training samples coupled with the complexity of the features that comprise the subject matter.  That combined with the large image size and the assumptions made by UMAP suggest that there is a lower bound on the number of training samples such that any elision whatsoever risks underfitting in the training procedure.

Future work will attempt to incorporate a predictive quantification of the impact of image size (and thus dimensionality) and feature complexity into the quality assessment.

\printbibliography{}

\end{document}